\documentclass{article}
\usepackage{ijcai22}

\usepackage{times}
\usepackage{soul}
\usepackage{url}
\usepackage[hidelinks]{hyperref}
\usepackage[utf8]{inputenc}
\usepackage[small]{caption}

\usepackage{subcaption}

\usepackage{graphicx}
\usepackage{amsmath}
\usepackage{amssymb}
\usepackage{amsthm}
\usepackage{booktabs}
\usepackage{algorithm}
\usepackage{algorithmic}
\usepackage[round]{natbib}

\title{Genetic Imitation Learning by Reward Extrapolation}

\author{
Boyuan Zheng$^1$
\and
Jianlong Zhou$^1$\And
Fang Chen$^1$
\affiliations
$^1$University of Technology Sydney\\
\emails
Boyuan.Zheng-1@student.uts.edu.au,
Jianlong.Zhou@uts.edu.au
}

\begin{document}

\maketitle
\begin{abstract}
  Imitation learning demonstrates remarkable performance in various domains. However, imitation learning is also constrained by many prerequisites. The research community has done intensive research to alleviate these constraints, such as adding the stochastic policy to avoid unseen states, eliminating the need for action labels, and learning from the suboptimal demonstrations. Inspired by the natural reproduction process, we proposed a method called GenIL that integrates the Genetic Algorithm with imitation learning. The involvement of the Genetic Algorithm improves the data efficiency by reproducing trajectories with various returns and assists the model in estimating more accurate and compact reward function parameters. We tested GenIL in both Atari and Mujoco domains, and the result shows that it successfully outperforms the previous extrapolation methods over extrapolation accuracy, robustness, and overall policy performance when input data is limited.
\end{abstract}

\section{Introduction}
Imitation learning, which leverages demonstration from other sources to reproduce the target behavior without handcrafting reward function, gets boosted with the help of recent advances in deep learning. Due to its ``reward function-free" characteristic, imitation learning plays a crucial role in problems where designing a reward function is impractical, such as robotic arm manipulation \citep{duanOneShotImitationLearning2017} and autonomous driving \citep{codevillaEndtoEndDrivingConditional2018}. However, the vast amount of existing research develops their methods, assuming that the input demonstrations are optimal or near-optimal. This assumption sets up an upper bound to the agent performance, and to some extent, it exacerbates the negative impact of suboptimal demonstrations on final performance \citep{zheng2021imitation}. The agent performance would converge to suboptimal when the input demonstrations are suboptimal. It is also a waste as the suboptimal demonstrations are more abundant and accessible to collect than the optimal demonstrations. Specifically, as imitation learning is interdisciplinary, optimal demonstrations from experts who are working in other research fields could be expensive and impractical.

Fortunately, a number of recent research \citep{changLearningSearchBetter2015,brownExtrapolatingSuboptimalDemonstrations2019,ding2019goal} have investigated and attempted to use the suboptimal demonstration to obtain better-than-demonstrator performance. Adversarial structured imitation learning and extrapolation are two mainstream approaches to address this problem. The state-of-the-art adversarial method GAIL \citep{hoGenerativeAdversarialImitation2016} presents considerable robustness on suboptimal input. Following methods like RAIL \citep{Zuo2020offpolicyRAIL} developed a more mature framework to better leverage suboptimal demonstration, but shortcomings like fragile structure are also inherent from the adversarial structure.
On the other hand, the research community has widely investigated another research direction: doing extrapolation induces better-than-demonstrator behavior. Besides the suboptimal demonstrations, the input commonly involves extra information, such as human preferences or the different levels of noise. The extra information facilitates the model to establish a relationship between the suboptimal demonstrations and the extra information and deduces the policy close to the optimal. The extrapolation achieves considerable performance in both Atari and Mujoco domains, but the extra information is commonly laborious. 

In this work, we propose an imitation learning method called Genetic Imitation Learning(GenIL). It advocates the Imitation from Observation(IfO) setting, integrates genetic algorithm concepts with imitation learning, and achieves more compact extrapolation using suboptimal demonstration. Unlike prior work, GenIL gets rid of the laborious procedure and leverages the idea from the genetic algorithm. By applying crossover and mutation on the input trajectories, GenIL generates new trajectories with multiple levels of quality. These newly generated trajectories serve as extra information to assist the extrapolation process. We compare our method with other related approaches on various Atari and Mujoco domains. The result shows that GenIL could achieve data efficiency and extrapolate more accurate and compact rewards for unseen trajectories than existing imitation learning algorithms. As for the overall policy performance, GenIL also presents competitive performance and outperforms previous methods.

The contributions of our work include:
\begin{itemize}
\item {We propose a better reward extrapolation model GenIL that generate a more compact and stable extrapolation to estimate the reward parameter by aggregating genetic algorithm and imitation learning.}
\item {GenIL makes better use of the suboptimal data and only uses two trajectories with diverse performance as input to train the model.}
\item {
We present a series of experiments and show that GenIL could obtain a meaningful policy from the suboptimal demonstration and outperform previous approaches.}
\end{itemize}

\section{Related Work}
This section briefly summarizes the recent advances in related topics learning from suboptimal demonstration, Imitation from Observation, reward extrapolation and genetic algorithm. 

Imitation learning aims to reproduce the target behavior with the help of demonstrations, it plays a crucial role in training autonomous agents. Typically, imitation learning could be categorized into behavior cloning \citep{bainFrameworkBehaviouralCloning1999} and inverse reinforcement learning \citep{russelllearning1998}. These two classes are distinguished by whether to restore the reward function. Commonly, they assume the input demonstration is optimal (or nearly optimal) and take state-action pairs batched from the optimal demonstrations as input.
As the research on imitation learning continues to deepen, researchers have begun to challenge these assumptions, including the challenge of the necessity of optimal demonstration. Especially after GAIL \citep{hoGenerativeAdversarialImitation2016} was proposed, the research community intensively investigated the feasibility of learning from suboptimal demonstrations. One of the commonly used techniques is integrating Hindsight Experience Replay(HER) \citep{andrychowicz2017hindsight} into GAIL, research such as HGAIL \citep{liu2019hindsight} and GoalGAIL \citep{ding2019goal} make use of the HER and set up new goals along the sub-optimal demonstrations to outperform the suboptimal demonstration. In RAIL \citep{Zuo2020offpolicyRAIL}, besides making an extension on HER, they also proposed a technique called hindsight copy which leverages both original demonstration and hindsight acquired demonstration to speed up the learning process.

With the emergence of new imitation learning algorithms, the restricted input form is also liberated. Until 2018, sequences of state-action pairs form the input data for imitation learning algorithms. The actions are commonly used as labels to train the model. However, in 2018, Liu et al. \citeyear{liuImitationObservationLearning2018} proposed a novel imitation paradigm called Imitation from Observation(IfO) that changed the previous tradition. Inspired by the nature of how humans and animals imitate, IfO eliminates the need for action labels and only utilizes the observations to train the model. IfO is working in a supervised learning manner, an encoder-decoder structure is adopted to extract features from the observations, and the distance between ground truth and predicted observation is measured to replace the conventional action label.
The presence of the IfO provides a solution for the scenarios where action labels are not available and enlarges the available training resource. Intensive research \citep{torabiImitationLearningVideo2019,zhu2021offpolicyifo,aytar2018playing} is attracted to leverage raw observation as input. For example, Sermanet et al. \citeyear{sermanetTimeContrastiveNetworksSelfSupervised2018} proposed a self-supervised IL method Time-Contrastive Network (TCN), that uses unlabeled multi-viewpoint video to assist the agent in learning its internal joint and invariant representation about the task.


\begin{figure}
  \centering
  \includegraphics[width=0.7\linewidth]{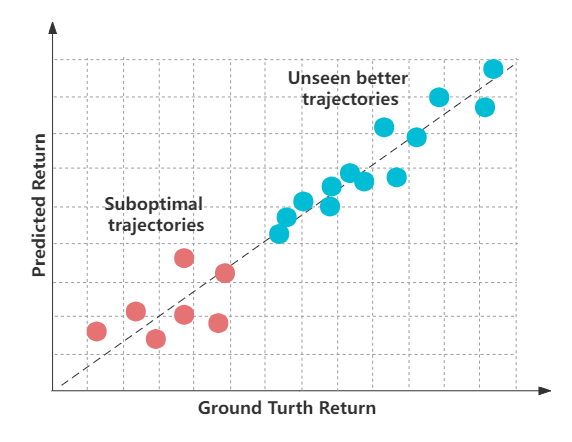}
  \caption{Graphical explanation about extrapolation. The red points are the suboptimal demonstrations used for training, and the blue points are the unseen trajectories.}
  \label{fig_ex}
\end{figure}

In addition to using the adversarial structure to address the suboptimal demonstration problem, the research community also explores other alternatives for this problem. Extrapolating rewards from the suboptimal demonstration is one of the reasonable solutions. Extrapolation means using additional information or reward signals to assist the model in establishing a connection between the suboptimal demonstration and intended goals. Figure \ref{fig_ex} provides a graphical explanation about extrapolation. Compared with adversarial structured imitation learning, extrapolation requires less environmental interaction and a more stable structure, but adding meaningful information could be laborious and expensive. Palan et al. \citeyear{palanLearningRewardFunctions2019} proposed DemPref that utilizes online preference from humans to overcome the demonstration quality problem in real-world robotic arm scenarios. Similarly, Brown et al. \citeyear{brownExtrapolatingSuboptimalDemonstrations2019} indicated that ranking the dataset in advance could help the agent infer better-than-demonstrator behaviors. The proposed T-REX inputs a gradually better demonstration dataset or manually ranked dataset for standard inverse reinforcement learning and outperforms the state-of-the-art methods on suboptimal scenarios. Their later approach, D-REX \citep{brownBetterthanDemonstratorImitationLearning2019}, eliminates the laborious procedure and achieves ranking by injecting different levels of noise. However, the noise is not always harmful to agent performance. Involving appropriate stochastic steps in trajectories could facilitate the generalization and prevent overfitting to the training dataset distribution, as shown in \citep{brownBetterthanDemonstratorImitationLearning2019}.
In this case, this paper uses stochastic policy more flexibly to assist in achieving better policy performance while implementing automated ranking.

\begin{figure*}[t]
      \centering
      \vspace{.3in}
      \includegraphics[width=0.95\linewidth]{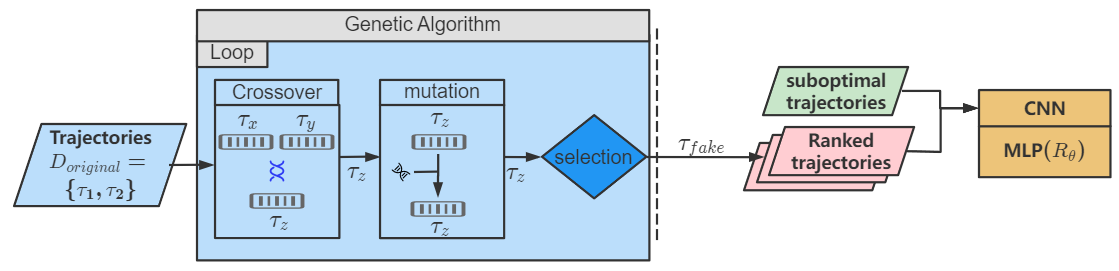}
      \vspace{.3in}
      \caption{A diagrammatic visualization of our method. The suboptimal demonstrations $\mathcal{D}$ firstly serve as the individual chromosomes in the genetic algorithm part to generate ``fake" trajectories $\tau_{fake}$, then together with the generated dataset to feed into the convolutional neural network. The features extracted by the convolution neural network are used to train the reward parameter $R_\theta$ represented by MLP.}
      \label{fig_network}
\end{figure*}

On the other hand, this paper also leverages the idea of Genetic Algorithm (GA). GA, an adaptive stochastic optimization algorithm, is one of the evolutionary algorithms \citep{karakativc2015survey}. GA was firstly proposed in \citep{de1975analysis}, and got further improved by \citep{holland1992adaptation,goldberg2006genetic} to solve the complex optimization problems. 
The research community developed several works in combining evolutionary algorithms with imitation learning in the fields of game strategy learning \citep{eliya2020evolutionary}, robot movement \citep{park2008imitation} and autonomous driving \citep{jalali2019optimal}. They leveraged evolutionary algorithms to either achieve offline dataset enrichment or optimize the network parameter.
GA process mimics how genes generate and evolve to achieve efficient searching and optimization. The genetic interchange is relatively sparse in the real world and occurs when producing new generations. GA maintains three basic biology principles: reproduction, natural selection and individual diversity \citep{darwin1859origin}, while bypassing other biological activities. It includes different problem features as a chromosome or genotype and iteratively reproduces new individuals by genetic operators. For instance, crossover interchanges the genes between two chromosomes and generates new offspring, and mutation mutates genetic snippets on a chromosome.
One individual represents one possible solution and participates in the natural selection process, evaluated by the fitness function. The fitness function filters out the weak individual and achieves optimization. In addition to having an appropriate fitness function, in order to efficiently obtain the optimal solution, individual diversity is also a crucial influencing factor. When individuals are scattered in their features, GA procedure could be more effective in finding the optimal global solution. 

In this work, we introduce GA into imitation learning and propose a method called GenIL. GenIL leverages the genetic operators,crossover and mutation, on the suboptimal trajectories and generates new samples with various levels of returns for extrapolating reward function parameters.

\section{Problem definition}

\begin{algorithm*}[tb]
\caption{GenIL}
\label{alg:algorithm}
\textbf{Input}: Demonstrations $\mathcal{D}$, number of offspring K, mutation rate $p_{mut}$, crossover rate $p_{crx}$\\
\textbf{Output}: Estimated reward parameter $R_\theta$

\begin{algorithmic}[1] 
\STATE Initialize reward parameter $R_\theta$ randomly.
\STATE relabel demonstrations with the initial rank$\mathcal{D} = \{( \tau_1,Rank_1), (\tau_2,Rank_2)\} $.
\WHILE{\textit{number of (offspring)}   $<$ K}
\STATE $\tau_x, Rank_x, \tau_y, Rank_y$ = SampleParents($\mathcal{D}$)
\STATE offspring $\tau_{fk}, Rank_fk$ = Crossover($p_{crx}$,$\tau_x, Rank_x, \tau_y, Rank_y$)
\STATE offspring $\tau_{fk}, Rank_fk$ = Mutation($p_{mut}$, $\tau_{fk}, Rank_fk$)
\IF {$Rank_fk$ satisfy selection rule}
\STATE Add offspring $(\tau_{fk},Rank_fk)$ to $\mathcal{D}_{rank}$
\STATE $\mathcal{D} \to \mathcal{D} \cup \mathcal{D}_{rank}$
\ENDIF
\ENDWHILE
\STATE Run T-REX to obtain $R_\theta$
\STATE Obtain policy $\pi_\theta$ by reinforcement learning using  reward parameter $R_\theta$.
\end{algorithmic}
\end{algorithm*}

In this work, we assume that GenIL is modeled under the Markov Decision Process (MDP) framework. MDP is the process satisfying the property that the next state $s_{t+1}$ only depends on the current state $s_t$ at any time $t$. Typically, a MDP is defined as a tuple ($\mathcal{S}$,$\mathcal{A}$,$\mathcal{P}$,$\gamma$,$\mathcal{D}$,$\mathcal{R}$), where $\mathcal{S}$ is the finite set of states, $\mathcal{A}$ is the corresponding set of actions, $\mathcal{P}$ is the set of state transition probabilities and the successor states $s_{t+1}$ is drawn from this transition model, i.e. $s_{t+1}=P(\cdot|s_t,a_t)$, $\gamma \in [1,0)$ is the discount factor, $\mathcal{D}$ is the set of demonstration trajectories $\mathcal{D}_{n} = \{\tau_1,\tau_2,...,\tau_n\}$ and $\mathcal{R}$ is the reward function $\mathcal{S} \mapsto \mathcal{R}$. The expected cumulative reward of a trajectory $\tau_x $ is $G (\tau_x) = \sum^{T}_{t=0}\gamma^{t}R(s^t)$.

Similar to other imitation learning methods, we assume the optimal reward function is not accessible. The input demonstration $\mathcal{D}_{original}$ consists of two trajectories: one good (or near-optimal) trajectory $\tau_{good}$ and one suboptimal demonstration $\tau_{bad}$. Here we wish to approximate a reward function parameter $\theta$, which reflects the intended optimal policy as close as possible. To represent the goodness of the trajectories, we propose to use a sequence of identical number to replace the original step reward sequence, i.e. $R'(S) = \{rank\}^m$, where $\sum R'_{\tau_{good}}(S) > \sum R'_{\tau_{bad}}(S)$. 
In previous work like T-REX \citep{brownExtrapolatingSuboptimalDemonstrations2019}, ranks are manually assigned by the expert. The ranking serves as the training label via supervised learning. This work inherits the previous setting while eliminating the laborious ranking procedure for human experts, the ranking is generated by genetic algorithm.
By implementing crossover and mutation steps, the original trajectories mix up and generate fake trajectories $\tau_{fake}$ with various numbers as their fake step reward $\sum R'_{\tau_{good}}(S) > \sum R'_{\tau_{fake}}(S) > \sum R'_{\tau_{bad}}(S)$. According to the accumulated fake step reward, these fake trajectories are classified into different ranks, and form ranked input together with the original dataset. The network trained on this combined dataset is expected to predict a more significant return on the trajectory that contains more parts from the good demonstration, i.e. $\mathbb{E}_{\pi}[\sum \gamma R_{\theta}(\tau_a)] > \mathbb{E}_{\pi}[\sum \gamma R_{\theta}(\tau_b)]$ if $R'(\tau_a)>R'(\tau_b)$.


\section{Method}

As mentioned in the introduction section, GenIL advocates the IfO setting while making use of the genetic algorithm. We hypothesize that the GA could achieve a meaningful ranking for reward inference. Like other IfO methods, GenIL uses state information to train the model. Given a demonstration dataset that is comprised of two trajectories $\mathcal{D}_{original} = \{\tau_1,\tau_2\}$ with diverse performance, the target is to learn a reward inference model trained by a neural network to distinguish the better trajectory from the unseen trajectory pairs, i.e., $\sum_{s\in\tau_x} \mathcal{R}_\theta(s) < \sum_{s\in\tau_y} \mathcal{R}_\theta(s)$ if $\sum R'(\tau_x) > \sum R'(\tau_y)$. It is proved that involving extra information is necessary to recover the correct reward function \citep{brownBetterthanDemonstratorImitationLearning2019}. Many prior works \citep{palanLearningRewardFunctions2019,brownExtrapolatingSuboptimalDemonstrations2019,castro2019inverse} either leverage human preference or inject noise to provide ranking as the extra information. Appropriately ranking the instances could be regarded as an example of the ordinal regression problem, leading to less ambiguity \citep{castro2019inverse,brownBetterthanDemonstratorImitationLearning2019}. This work follows the prior framework and proposes a novel learning paradigm GenIL, that leverages genetic algorithms as the extra information to achieve ranking and assist the reward inference. Figure \ref{fig_network} demonstrates the working process of GenIL.

GenIL consists of two components. (1) The genetic algorithm part operates crossover and mutation steps on two trajectories randomly sampled from the dataset, and outputs a "fake" trajectory, i.e., 
$
\tau_{fake} = \{s_i | s_i \in \tau_1 \cup \tau_2 \cup S_{mut} \}\text{ and }
$
\begin{equation}
\resizebox{.99\linewidth}{!}{
$
\sum_{s_i \in \tau_{fake}} R'(s_i) = \sum_{s_j \in S_j \subset \tau_{1}} R'(s_j) +\sum_{s_k \in S_k \subset \tau_{2}} R'(s_k) + \sum_{s_n \in S_{mut}} R'(s_n)
\nonumber
$
}
\end{equation}
\noindent
where $S_{mut}$ is the available mutation sample set, the states in the mutation set is batched from all visited states. Meanwhile, in order to maintain the randomness and continuity between states, we set the crossover step size in a random interval that is less than 10, and the mutation step will provide a random rank on random states. As for the selection part, we implement simple selection criteria: If the generated offspring's average rank lies in the predefined interval, then this offspring is added to the corresponding ranking dataset.
After selection steps, the ranking dataset is combined with the original dataset as the ranked dataset, i.e., $D_{ranked} = D_{original} + D_{fake}$. In the experiment, we treat the number of generated offspring as a hyperparameter and set its value as 12 based on trial and error. Implementing GA on a generated dataset could make better use of the data and outperform prior methods on data efficiency. (2) IRL part learns a reward inference model from the generated ranked dataset. In order to extract features from frames, we first implement a four-layer convolutional neural network, followed by a multilayer perceptron(MLP) to reference reward. The network takes trajectories as input and outputs an expected reward of the given trajectory. Similar to prior work, the model is trained in a supervised learning fashion with the pairwise ranking loss adapted from  \citep{brownExtrapolatingSuboptimalDemonstrations2019}:  
\[
\mathcal{L}(\theta) \approx -\sum_{\tau_i,\tau_j} \log \frac{\exp\sum_{s\in\tau_j}R_\theta(s)}{\exp\sum_{s\in\tau_i}R_\theta(s)+\exp\sum_{s\in\tau_j}R_\theta(s)}
\]
where $\sum_{s \in \tau_{j}} R'(s) >  \sum_{s \in \tau_{i}} R'(s)$. This loss function is in the form of Plackett-Luce model \citep{plackett1975analysis,luce2012individual}, which demonstrates effectiveness in learning a model from ranked instance by neural network \citep{christiano2017deep,mohlin2020learning}.
The obtained reward function parameter $\theta$ could be further used by other reinforcement learning algorithms to obtain the policy. The algorithmic description of GenIL is presented in Algorithm \ref{alg:algorithm}.

\section{Experiments}
The algorithm GenIL proposed above implements a basic genetic algorithm under imitation from observation paradigms. We hypothesize that the implementation of the genetic algorithm could make better use of the suboptimal data and achieve competitive performance while achieving data efficiency and robustness. This section introduces the experimental design to validate our hypothesis and discuss the results.

\begin{figure*}[t]
      \centering
      \vspace{.3in}
      \includegraphics[width=0.95\textwidth]{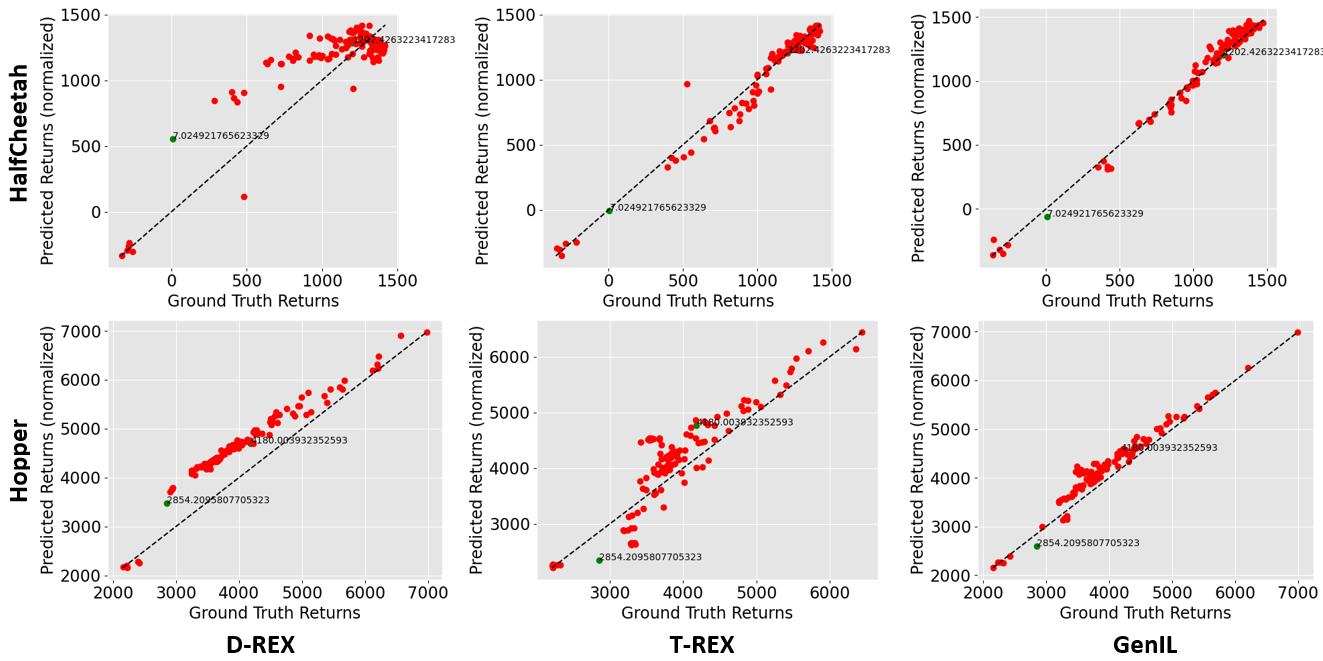}
      \vspace{.3in}
      \caption{Extrapolation graphical comparison. The green points are the data used for training, and the red points are the unseen trajectories for prediction. The predicted and ground truth returns are normalized into the same range for comparison.}
      \label{fig_compare}
\end{figure*}

\subsection{Setup}
We evaluate GenIL on five prevalent benchmarks on GPU NVIDIA Quadro RTX 5000. Two robotic tasks by the Mujoco simulator \citep{todorov2012mujoco} within OpenAI Gym \citep{brockman2016openai}: Hopper, HalfCheetah, and three Atari tasks: Breakout, Beamrider, SpaceInvaders. 

We implement a three-layer neural network to represent the reward parameter with 256 neurons and the ReLU activation function for each layer. For Mujoco tasks, we directly use the state information provided by the simulator to train the model. While in Atari tasks, we use the raw frames as input so that we add a four-layer convolutional neural network in front of the fully connected layers with a leaky-ReLU activation function for each layer. To avoid the ``causal confusion" problem \citep{dehaanCausalConfusionImitation2019}, which establishes a wrong relationship with the irrelevant features, we mask the indicators part such as score and remaining life.

Referring to the recent approach, we use the proximal policy optimization(PPO) \citep{schulman2017proximal} implemented by OpenAI baselines \citep{dhariwal2017openai} with default parameters and reward function for generating the dataset. The PPO training is checkpointed every 20 training steps. To represent the difference in agent performance, we recover two agents, one from a checkpoint in the early stage and another close to the end. For Atari tasks, we use checkpoints 400 and 1000, while under the Mujoco domain, we choose checkpoints 40 and 220 as the input. The interactions between the recovered agents and the task environment are recorded as trajectories. To ensure evaluation fairness, the initial trajectories are recorded and serve as the input for other baseline methods. Together with the GA-generated trajectories, these trajectories are sub-sampled into about 5000 snippets with various lengths ranging from 100 to 300 and feed into the network. As for the hyperparameters for GA, we set the fixed mutation rate and crossover rate to 0.05 and 0.9, respectively. To obtain the policy from the extrapolated reward parameter, we train the policy over 480 training steps in the Mujoco domain, while in the Atari domain, we evaluate the policy with 3000 training steps. 

We compared the average performance between GenIL and other methods such as BC, T-REX and D-REX from 20 trials that trained five models in each trial with various random seeds. Specifically, as there is relatively little work on using extrapolation to learn reward functions for imitation learning, we compared the extrapolation performance with T-REX on aspects like accuracy and robustness.

\subsection{Result}

We hypothesize that GenIL could achieve better extrapolation with the help of the basic genetic algorithm over two aspects: (1) The extrapolation accuracy on the unseen trajectories, we use the ratio of the normalized predicted return and ground truth, i.e., $average(\frac{R_\theta(\tau_t)}{R^*(\tau_t)})$, to measure it. (2) The overall performance of the trained agents on the target domain, which is measured by the ground truth return obtained from simulation.

Figure \ref{fig_compare} demonstrates the extrapolation comparison between the proposed GenIL and the previous extrapolation algorithms T-REX and D-REX under the Mujoco domain. The green points represent two training trajectories generated from deviated checkpoints. The unseen trajectories that could be better or even worse are depicted as the red points, and we aim to extrapolate compact predictive rewards for these unseen trajectories. The x-axis represents the ground truth returns of various demonstration levels, while the y-axis is the return predicted by the extrapolation model. The expected and ground truth returns are scaled into the same range for convenience. The dashed line is an indicator to demonstrate the deviation between predicted and ground truth returns.

\begin{table*}[ht]
\caption{The policy performance comparison between GenIL and previous method T-REX. The value in this table is obtained from 20 trials with different random seeds. }
\label{table_pp}
\centering
\begin{tabular}{@{}lllllllll@{}}
\toprule
\multicolumn{1}{c}{} & \multicolumn{2}{c}{\textbf{GenIL}} & \multicolumn{2}{c}{\textbf{T-REX}} & \multicolumn{2}{c}{\textbf{D-REX}} & \multicolumn{2}{c}{\textbf{BC}} \\
\multicolumn{1}{c}{\textbf{Tasks}} & \multicolumn{1}{c}{\textbf{Avg}} & \multicolumn{1}{c}{\textbf{Std}} & \multicolumn{1}{c}{\textbf{Avg}} & \multicolumn{1}{c}{\textbf{Std}} & \multicolumn{1}{c}{\textbf{Avg}} & \multicolumn{1}{c}{\textbf{Std}} & \multicolumn{1}{c}{\textbf{Avg}} & \multicolumn{1}{c}{\textbf{Std}} \\ \midrule
\textbf{HalfCheetah} & \textbf{1765.9} & 383.7 & 1397.3 & 408.7 & 378.4 & 542.1 & -364.3 & 2.0 \\
\textbf{Hopper} & \textbf{1693.0} & 657.3 & 1342.1 & 473.8 & 482.1 & 315.2 & 800.6 & 30.9 \\ \midrule
\textbf{Beamrider} & \textbf{927.6} & 647.2 & 825.4 & 265.5 & 843.7 & 176.3 & 477.2 & 155.5 \\
\textbf{Breakout} & \textbf{17.5} & 6.3 & 14.1 & 5.9 & 11.8 & 2.9 & 2.1 & 4.1 \\
\textbf{Spaceinvaders} & \textbf{478.1} & 192.5 & 364.3 & 151.2 & 299.3 & 87.6 & 127.3 & 85.4 \\ \bottomrule
\end{tabular}
\end{table*}

From Figure \ref{fig_compare}, we can see all three extrapolation methods perform better in Halfcheetah compared to Hopper, and their predictions are closer to the actual value. In HalfCheetah, the unseen trajectories are intensively located at the top-right part of the plot, which indicates the original PPO policy gradually converges at the true reward of 1500. In contrast, in Hopper, most of the unseen trajectories are concentrated around the true reward of 4000. As for the comparison between methods, GenIL presents smaller vertical deviations in the prediction for data points with similar ground-truth rewards, which means GenIL takes advantage of the generated fake trajectories and makes a more compact and robust prediction for unseen trajectories.  

\begin{figure}[t]
      \centering
      \vspace{.3in}
      \includegraphics[width=0.45\textwidth]{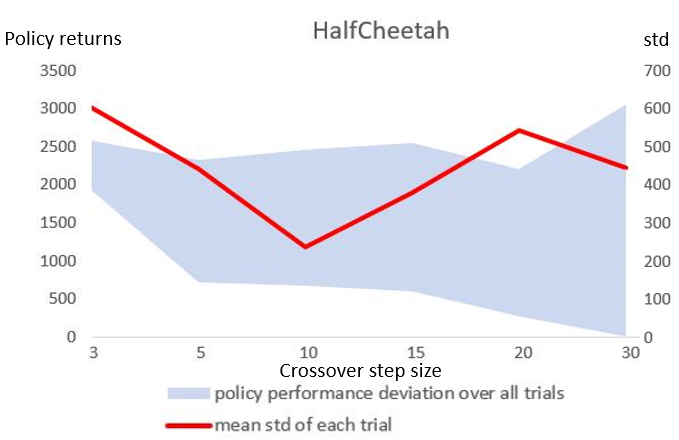}
      \vspace{.3in}
      \caption{Policy performance change with respect to crossover step size. The blue area reflects policy performance over from 10 trials for each crossover step size under various random seeds. The red line is the performance's average standard deviation from five trained models of each trial using the same crossover step size.}
      \label{fig_hyper}
\end{figure}

Table \ref{table_pp} compared the policy performance between GenIL and previous methods. The value in this table is obtained from 20 trials with different random seeds. From Table \ref{table_pp}, we can conclude that GenIL outperforms the previous method in all tasks. The obtained policy maintains a competitive performance in different tasks over previous methods using limited demonstrations. The amount of inputs seems to have less influence on GenIL and T-REX. In contrast, the demonstration quality significantly restricts BC and D-REX. BC fails to learn meaningful policy in most of the tasks, the demonstrations set up an upper bound for BC's performance. Compared to T-REX and GenIL, the standard deviation of D-REX's performance is the smallest, which indicates that D-REX is more stable than T-REX and GenIL. The embedding BC process generates sufficient demonstrations in each rank so that the produced policy can be more stable. However, the BC process limited its performance. This could explain why BC and D-REX perform poorly in some tasks simultaneously.
Comparing the performance between two classes of tasks, GenIL's performance deviation in the discrete Atari tasks is much larger than the deviation in continuous Mujoco tasks. 

For the Mujoco tasks, BC fails to learn meaningful policy in HalfCheetah, as the suboptimal demonstrations might not be sufficient to establish an executable mapping between states and actions. At the same time, BC's performance is also limited by the ``compounding error" \citep{rossEfficientReductionsImitation2010} so that the learned policy could not recover from the unseen states. D-REX is also influenced by the BC limitation and fails to extrapolate good reward parameters. But the effect of the numerous BC-generated trajectories indicates that the performance deviation significantly depends on the number of training samples, even if the samples are far suboptimal. 
T-REX achieves good policy performance in HalfCheetah and Hopper and slightly outperforms the suboptimal demonstration, but the deviation of the policy performance is the largest overall evaluated method. From Table \ref{table_pp}, we can see that GenIL also achieves better performance in HalfCheetah and Hopper compared with other evaluated methods, especially in HalfCheetah, GenIL surpasses the demonstrations and T-REX by a large margin, which is consistent with Figure \ref{fig_compare} that GenIL makes more compact and stable reward predictions of the unseen data. 

Throughout experiments, we find that the intrinsic parameters of genetic algorithms could influence learning performance. For example, when the difference between the initial input trajectories is significant, increasing the number of ranks could lead to more accurate extrapolation, better policy performance, and smaller variance. However, when the difference between the initial input trajectories is small, fewer ranks could obtain good performance as well as save time in finding the candidates for the end ranks. The number of generated fake trajectories also plays a critical role in extrapolation. The predictions of the unseen data clustered more compact when more fake offspring are used in training, which presents similar characteristics as the D-REX. 
Another important parameter in GenIL could be the ratio between the crossover step size and the sub-sampling size. To ensure the effectiveness of crossover, we set the crossover step size to less than 10, while the sub-sampling size is about 200. In this case, there is a sufficient number of small snippets from the initial trajectories in the fake trajectories after crossover, and the number of small snippets from the good initial trajectory, in fact, becomes the indicator of the strength of the fake trajectories. Take the HalfCheetah domain as an example, we evaluated the policy performance under different crossover step sizes (see Figure \ref{fig_hyper}).

In Figure \ref{fig_hyper}, the x-axis is the different crossover step sizes, the left y-axis is the policy performance, and the right y-axis reflects the standard deviation over five models of each trial.
According to Figure \ref{fig_hyper}, we can see the crossover step size does not have an obvious influence on the upper bound of the policy performance, but the consistency of the policy performance is significantly impacted. When the crossover step size is too large compared to the subsampling size, the crossover effect is significantly reduced, and the differences between ranks become more blurred. As reflected in the policy performance and the standard deviation of each trial, the trained agents vary more significantly and behave more unstable when the crossover step size is too large. 
On the other hand, when the crossover step size is too small, the policy performance between trials becomes more consistent, but the standard deviation of each trial becomes larger. We suspect the reason for this problem is that the dynamic transition consistency cannot preserve well, which makes the fake trajectories fragmented.

\section{Limitations}
There are still some challenges in making better use of genetic algorithms in imitation learning. An important issue is the need for more dynamic hyperparameter selection. The genetic algorithm process introduces more hyperparameters for imitation learning.
In this work, the hyperparameters are selected through trial and error. Future work is needed to achieve better optimization.
Dynamic hyperparameter selection prevents dense crossovers and mutations from fragmenting and randomizing the generated trajectories. How to balance the generalization and fragmentation caused by genetic algorithms in imitation learning could be an interesting topic for further research. Since GenIL uses the same ranking for all observations in a trajectory, this may not make much sense. Future research could use statistics such as observational visit frequency or action frequency to differentiate segments and weigh more meaningful parts. More advanced selection rules and learning paradigms in GenIL can also be used in future work to improve inference and overall performance.

\section{Conclusion}
In this paper, we presented GenIL, a reward extrapolation method based on the combination of genetic algorithms and inverse reinforcement learning. GenIL leveraged crossover and mutation to reproduce the demonstration improvement process from far suboptimal to better and inferred the reward function parameter through IRL from the resulting ranking dataset. This extrapolation technique allowed us to make better use of the suboptimal demonstrations and eliminate the laborious preference process for experts. From the experiments, we have shown that GenIL has not only small fluctuations in extrapolation but also better overall policy performance than previous methods. 

\newpage
\clearpage
\bibliographystyle{named}
\bibliography{ijcai22}

\begin{thebibliography}{}

\bibitem[\protect\citeauthoryear{Andrychowicz \bgroup \em et al.\egroup
  }{2017}]{andrychowicz2017hindsight}
Marcin Andrychowicz, Filip Wolski, Alex Ray, Jonas Schneider, Rachel Fong,
  Peter Welinder, Bob McGrew, Josh Tobin, OpenAI~Pieter Abbeel, and Wojciech
  Zaremba.
\newblock Hindsight experience replay.
\newblock In {\em Advances in neural information processing systems}, pages
  5048--5058, 2017.

\bibitem[\protect\citeauthoryear{Aytar \bgroup \em et al.\egroup
  }{2018}]{aytar2018playing}
Yusuf Aytar, Tobias Pfaff, David Budden, Tom~Le Paine, Ziyu Wang, and Nando
  de~Freitas.
\newblock Playing hard exploration games by watching youtube.
\newblock {\em arXiv preprint arXiv:1805.11592}, 2018.

\bibitem[\protect\citeauthoryear{Bain and
  Sammut}{1999}]{bainFrameworkBehaviouralCloning1999}
Michael Bain and Claude Sammut.
\newblock A framework for behavioural cloning.
\newblock In {\em Machine {Intelligence} 15}, pages 103--129. Oxford University
  Press, 1999.

\bibitem[\protect\citeauthoryear{Brockman \bgroup \em et al.\egroup
  }{2016}]{brockman2016openai}
Greg Brockman, Vicki Cheung, Ludwig Pettersson, Jonas Schneider, John Schulman,
  Jie Tang, and Wojciech Zaremba.
\newblock Openai gym.
\newblock {\em arXiv preprint arXiv:1606.01540}, 2016.

\bibitem[\protect\citeauthoryear{Brown \bgroup \em et al.\egroup
  }{2019}]{brownExtrapolatingSuboptimalDemonstrations2019}
Daniel Brown, Wonjoon Goo, Prabhat Nagarajan, and Scott Niekum.
\newblock Extrapolating beyond suboptimal demonstrations via inverse
  reinforcement learning from observations.
\newblock In {\em International conference on machine learning}, pages
  783--792. PMLR, 2019.

\bibitem[\protect\citeauthoryear{Brown \bgroup \em et al.\egroup
  }{2020}]{brownBetterthanDemonstratorImitationLearning2019}
Daniel~S Brown, Wonjoon Goo, and Scott Niekum.
\newblock Better-than-demonstrator imitation learning via automatically-ranked
  demonstrations.
\newblock In {\em Conference on robot learning}, pages 330--359. PMLR, 2020.

\bibitem[\protect\citeauthoryear{Castro \bgroup \em et al.\egroup
  }{2019}]{castro2019inverse}
Pablo~Samuel Castro, Shijian Li, and Daqing Zhang.
\newblock Inverse reinforcement learning with multiple ranked experts.
\newblock {\em arXiv preprint arXiv:1907.13411}, 2019.

\bibitem[\protect\citeauthoryear{Chang \bgroup \em et al.\egroup
  }{2015}]{changLearningSearchBetter2015}
Kai-Wei Chang, Akshay Krishnamurthy, Alekh Agarwal, Hal Daum{\'e}~III, and John
  Langford.
\newblock Learning to search better than your teacher.
\newblock In {\em International Conference on Machine Learning}, pages
  2058--2066. PMLR, 2015.

\bibitem[\protect\citeauthoryear{Christiano \bgroup \em et al.\egroup
  }{2017}]{christiano2017deep}
Paul Christiano, Jan Leike, Tom~B Brown, Miljan Martic, Shane Legg, and Dario
  Amodei.
\newblock Deep reinforcement learning from human preferences.
\newblock {\em arXiv preprint arXiv:1706.03741}, 2017.

\bibitem[\protect\citeauthoryear{Codevilla \bgroup \em et al.\egroup
  }{2018}]{codevillaEndtoEndDrivingConditional2018}
Felipe Codevilla, Matthias M{\"u}ller, Antonio L{\'o}pez, Vladlen Koltun, and
  Alexey Dosovitskiy.
\newblock End-to-end driving via conditional imitation learning.
\newblock In {\em 2018 IEEE International Conference on Robotics and Automation
  (ICRA)}, pages 4693--4700. IEEE, 2018.

\bibitem[\protect\citeauthoryear{Darwin}{1859}]{darwin1859origin}
Charles Darwin.
\newblock {\em The origin of species by means of natural selection}.
\newblock Pub One Info, 1859.

\bibitem[\protect\citeauthoryear{de Haan \bgroup \em et al.\egroup
  }{2019}]{dehaanCausalConfusionImitation2019}
Pim de~Haan, Dinesh Jayaraman, and Sergey Levine.
\newblock Causal confusion in imitation learning.
\newblock {\em Advances in Neural Information Processing Systems},
  32:11698--11709, 2019.

\bibitem[\protect\citeauthoryear{De~Jong}{1975}]{de1975analysis}
Kenneth~Alan De~Jong.
\newblock {\em An analysis of the behavior of a class of genetic adaptive
  systems.}
\newblock University of Michigan, 1975.

\bibitem[\protect\citeauthoryear{Dhariwal \bgroup \em et al.\egroup
  }{2017}]{dhariwal2017openai}
Prafulla Dhariwal, Christopher Hesse, Oleg Klimov, Alex Nichol, Matthias
  Plappert, Alec Radford, John Schulman, Szymon Sidor, Yuhuai Wu, and Peter
  Zhokhov.
\newblock Openai baselines, 2017.

\bibitem[\protect\citeauthoryear{Ding \bgroup \em et al.\egroup
  }{2019}]{ding2019goal}
Yiming Ding, Carlos Florensa, Mariano Phielipp, and Pieter Abbeel.
\newblock Goal-conditioned imitation learning.
\newblock {\em arXiv preprint arXiv:1906.05838}, 2019.

\bibitem[\protect\citeauthoryear{Duan \bgroup \em et al.\egroup
  }{2017}]{duanOneShotImitationLearning2017}
Yan Duan, Marcin Andrychowicz, Bradly~C Stadie, Jonathan Ho, Jonas Schneider,
  Ilya Sutskever, Pieter Abbeel, and Wojciech Zaremba.
\newblock One-shot imitation learning.
\newblock {\em arXiv preprint arXiv:1703.07326}, 2017.

\bibitem[\protect\citeauthoryear{Eliya and
  Herrmann}{2020}]{eliya2020evolutionary}
Roy Eliya and J~Michael Herrmann.
\newblock Evolutionary selective imitation: Interpretable agents by imitation
  learning without a demonstrator.
\newblock {\em arXiv preprint arXiv:2009.08403}, 2020.

\bibitem[\protect\citeauthoryear{Goldberg}{2006}]{goldberg2006genetic}
David~E Goldberg.
\newblock {\em Genetic algorithms}.
\newblock Pearson Education India, 2006.

\bibitem[\protect\citeauthoryear{Ho and
  Ermon}{2016}]{hoGenerativeAdversarialImitation2016}
Jonathan Ho and Stefano Ermon.
\newblock Generative adversarial imitation learning.
\newblock {\em Advances in neural information processing systems},
  29:4565--4573, 2016.

\bibitem[\protect\citeauthoryear{Holland}{1992}]{holland1992adaptation}
John~H Holland.
\newblock {\em Adaptation in natural and artificial systems: an introductory
  analysis with applications to biology, control, and artificial intelligence}.
\newblock MIT press, 1992.

\bibitem[\protect\citeauthoryear{Jalali \bgroup \em et al.\egroup
  }{2019}]{jalali2019optimal}
Seyed Mohammad~Jafar Jalali, Parham~M Kebria, Abbas Khosravi, Khaled Saleh,
  Darius Nahavandi, and Saeid Nahavandi.
\newblock Optimal autonomous driving through deep imitation learning and
  neuroevolution.
\newblock In {\em 2019 IEEE International Conference on Systems, Man and
  Cybernetics (SMC)}, pages 1215--1220. IEEE, 2019.

\bibitem[\protect\citeauthoryear{Karakati{\v{c}} and
  Podgorelec}{2015}]{karakativc2015survey}
Sa{\v{s}}o Karakati{\v{c}} and Vili Podgorelec.
\newblock A survey of genetic algorithms for solving multi depot vehicle
  routing problem.
\newblock {\em Applied Soft Computing}, 27:519--532, 2015.

\bibitem[\protect\citeauthoryear{Liu \bgroup \em et al.\egroup
  }{2018}]{liuImitationObservationLearning2018}
YuXuan Liu, Abhishek Gupta, Pieter Abbeel, and Sergey Levine.
\newblock Imitation from observation: Learning to imitate behaviors from raw
  video via context translation.
\newblock In {\em 2018 IEEE International Conference on Robotics and Automation
  (ICRA)}, pages 1118--1125. IEEE, 2018.

\bibitem[\protect\citeauthoryear{Liu \bgroup \em et al.\egroup
  }{2019}]{liu2019hindsight}
Naijun Liu, Tao Lu, Yinghao Cai, Boyao Li, and Shuo Wang.
\newblock Hindsight generative adversarial imitation learning.
\newblock {\em arXiv preprint arXiv:1903.07854}, 2019.

\bibitem[\protect\citeauthoryear{Luce}{2012}]{luce2012individual}
R~Duncan Luce.
\newblock {\em Individual choice behavior: A theoretical analysis}.
\newblock Courier Corporation, 2012.

\bibitem[\protect\citeauthoryear{Mohlin \bgroup \em et al.\egroup
  }{2020}]{mohlin2020learning}
Erik Mohlin, Robert {\"O}stling, and Joseph Tao-yi Wang.
\newblock Learning by similarity-weighted imitation in winner-takes-all games.
\newblock {\em Games and Economic Behavior}, 120:225--245, 2020.

\bibitem[\protect\citeauthoryear{Palan \bgroup \em et al.\egroup
  }{2019}]{palanLearningRewardFunctions2019}
Malayandi Palan, Nicholas~C Landolfi, Gleb Shevchuk, and Dorsa Sadigh.
\newblock Learning reward functions by integrating human demonstrations and
  preferences.
\newblock {\em arXiv preprint arXiv:1906.08928}, 2019.

\bibitem[\protect\citeauthoryear{Park \bgroup \em et al.\egroup
  }{2008}]{park2008imitation}
GaLam Park, Syungkwon Ra, ChangHwan Kim, and JaeBok Song.
\newblock Imitation learning of robot movement using evolutionary algorithm.
\newblock {\em IFAC Proceedings Volumes}, 41(2):730--735, 2008.

\bibitem[\protect\citeauthoryear{Plackett}{1975}]{plackett1975analysis}
Robin~L Plackett.
\newblock The analysis of permutations.
\newblock {\em Journal of the Royal Statistical Society: Series C (Applied
  Statistics)}, 24(2):193--202, 1975.

\bibitem[\protect\citeauthoryear{Ross and
  Bagnell}{2010}]{rossEfficientReductionsImitation2010}
St{\'e}phane Ross and Drew Bagnell.
\newblock Efficient reductions for imitation learning.
\newblock In {\em Proceedings of the thirteenth international conference on
  artificial intelligence and statistics}, pages 661--668. JMLR Workshop and
  Conference Proceedings, 2010.

\bibitem[\protect\citeauthoryear{Russell}{1998}]{russelllearning1998}
Stuart Russell.
\newblock Learning agents for uncertain environments.
\newblock In {\em Proceedings of the eleventh annual conference on
  Computational learning theory}, pages 101--103, 1998.

\bibitem[\protect\citeauthoryear{Schulman \bgroup \em et al.\egroup
  }{2017}]{schulman2017proximal}
John Schulman, Filip Wolski, Prafulla Dhariwal, Alec Radford, and Oleg Klimov.
\newblock Proximal policy optimization algorithms.
\newblock {\em arXiv preprint arXiv:1707.06347}, 2017.

\bibitem[\protect\citeauthoryear{Sermanet \bgroup \em et al.\egroup
  }{2018}]{sermanetTimeContrastiveNetworksSelfSupervised2018}
Pierre Sermanet, Corey Lynch, Yevgen Chebotar, Jasmine Hsu, Eric Jang, Stefan
  Schaal, Sergey Levine, and Google Brain.
\newblock Time-contrastive networks: Self-supervised learning from video.
\newblock In {\em 2018 IEEE international conference on robotics and automation
  (ICRA)}, pages 1134--1141. IEEE, 2018.

\bibitem[\protect\citeauthoryear{Todorov \bgroup \em et al.\egroup
  }{2012}]{todorov2012mujoco}
Emanuel Todorov, Tom Erez, and Yuval Tassa.
\newblock Mujoco: A physics engine for model-based control.
\newblock In {\em 2012 IEEE/RSJ International Conference on Intelligent Robots
  and Systems}, pages 5026--5033. IEEE, 2012.

\bibitem[\protect\citeauthoryear{Torabi \bgroup \em et al.\egroup
  }{2019}]{torabiImitationLearningVideo2019}
Faraz Torabi, Garrett Warnell, and Peter Stone.
\newblock Imitation learning from video by leveraging proprioception.
\newblock {\em arXiv preprint arXiv:1905.09335}, 2019.

\bibitem[\protect\citeauthoryear{Zheng \bgroup \em et al.\egroup
  }{2021}]{zheng2021imitation}
Boyuan Zheng, Sunny Verma, Jianlong Zhou, Ivor Tsang, and Fang Chen.
\newblock Imitation learning: Progress, taxonomies and opportunities.
\newblock {\em arXiv preprint arXiv:2106.12177}, 2021.

\bibitem[\protect\citeauthoryear{Zhu \bgroup \em et al.\egroup
  }{2021}]{zhu2021offpolicyifo}
Zhuangdi Zhu, Kaixiang Lin, Bo~Dai, and Jiayu Zhou.
\newblock Off-policy imitation learning from observations.
\newblock {\em arXiv preprint arXiv:2102.13185}, 2021.

\bibitem[\protect\citeauthoryear{Zuo \bgroup \em et al.\egroup
  }{2020}]{Zuo2020offpolicyRAIL}
Guoyu Zuo, Qishen Zhao, Kexin Chen, Jiangeng Li, and Daoxiong Gong.
\newblock Off-policy adversarial imitation learning for robotic tasks with
  low-quality demonstrations.
\newblock {\em Applied Soft Computing}, 97:106795, 2020.

\end{thebibliography}

\end{document}